\documentclass[letterpaper]{article} %
\usepackage{aaai2026}  %
\usepackage{times}  %
\usepackage{helvet}  %
\usepackage{courier}  %
\usepackage[hyphens]{url}  %
\usepackage{graphicx} %
\usepackage{natbib}  %
\usepackage{caption} %
\usepackage{algorithm}
\usepackage{algorithmic}

\usepackage{newfloat}
\usepackage{listings}
\usepackage{booktabs}       %
\usepackage{amsfonts}       %
\usepackage{nicefrac}       %
\usepackage{microtype}      %
\usepackage{xcolor}         %
\usepackage{graphicx}
\usepackage{cleveref}
\usepackage{multirow}
\usepackage{marvosym} 
\DeclareCaptionStyle{ruled}{labelfont=normalfont,labelsep=colon,strut=off} %
\floatstyle{ruled}
\newfloat{listing}{tb}{lst}{}
\floatname{listing}{Listing}
\title{GenVidBench: A 6-Million Benchmark for AI-Generated Video Detection}
\renewcommand\footnotemark{}
\author{
Zhenliang Ni$^*$, 
Qiangyu Yan$^*$,
Mouxiao Huang, 
Tianning Yuan,  
Yehui Tang, \\
Hailin Hu$^{~\textrm{\Letter}}$,
Xinghao Chen$^{~\textrm{\Letter}}$,
Yunhe Wang$^{~\textrm{\Letter}}$\\
\thanks{$^*$~Equal contribution.}
\thanks{$^{~\textrm{\Letter}}$~Corresponding authors.}
}
\affiliations{
    Huawei Noah's Ark Lab\\
    \{nizhenliang2, qiangyu.yan, hailin.hu, xinghao.chen, yunhe.wang\}@huawei.com 
}

\usepackage{bibentry}

\begin{document}

\maketitle

\begin{abstract}
    The rapid advancement of video generation models has made it increasingly challenging to distinguish AI-generated videos from real ones. This issue underscores the urgent need for effective AI-generated video detectors to prevent the dissemination of false information via such videos. However, the development of high-performance AI-generated video detectors is currently impeded by the lack of large-scale, high-quality datasets specifically designed for generative video detection.
    To this end, we introduce GenVidBench, a challenging AI-generated video detection dataset with several key advantages: 
    1) Large-scale video collection: The dataset contains 6.78 million videos and is currently the largest dataset for AI-generated video detection.
    2) Cross-Source and Cross-Generator: The cross-source generation reduces the interference of video content on the detection. The cross-generator ensures diversity in video attributes between the training and test sets, preventing them from being overly similar.
    3) State-of-the-Art Video Generators: The dataset includes videos from 11 state-of-the-art AI video generators, ensuring that it covers the latest advancements in the field of video generation. 
    These generators ensure that the datasets are not only large in scale but also diverse, aiding in the development of generalized and effective detection models. 
    Additionally, we present extensive experimental results with advanced video classification models. With GenVidBench, researchers can efficiently develop and evaluate AI-generated video detection models.
\end{abstract}
\begin{links}
    \link{Code and Datasets}{https://genvidbench.github.io/}
\end{links}

\section{Introduction}
In recent years, video generation models like Sora have seen remarkable advancements~\cite{sora,svd,musev}, leading to a significant enhancement in the quality of AI-generated videos. The line between realistically generated videos and real videos has become increasingly blurred, posing challenges such as the spread of misinformation, damage to personal and corporate reputations, and an escalation in cybersecurity threats~\cite{aigdet, decof}. To address these risks, there is an urgent demand for the development of AI-generated video detectors that can accurately identify and differentiate between real and fake videos. However, the development and evaluation of such detectors are still hindered by the lack of large-scale, challenging datasets. To this end, we present GenVidBench, a comprehensive and challenging dataset for the development of AI-generated video detectors.

\begin{figure}[t] 
    \centering 
    \includegraphics[width=\columnwidth]{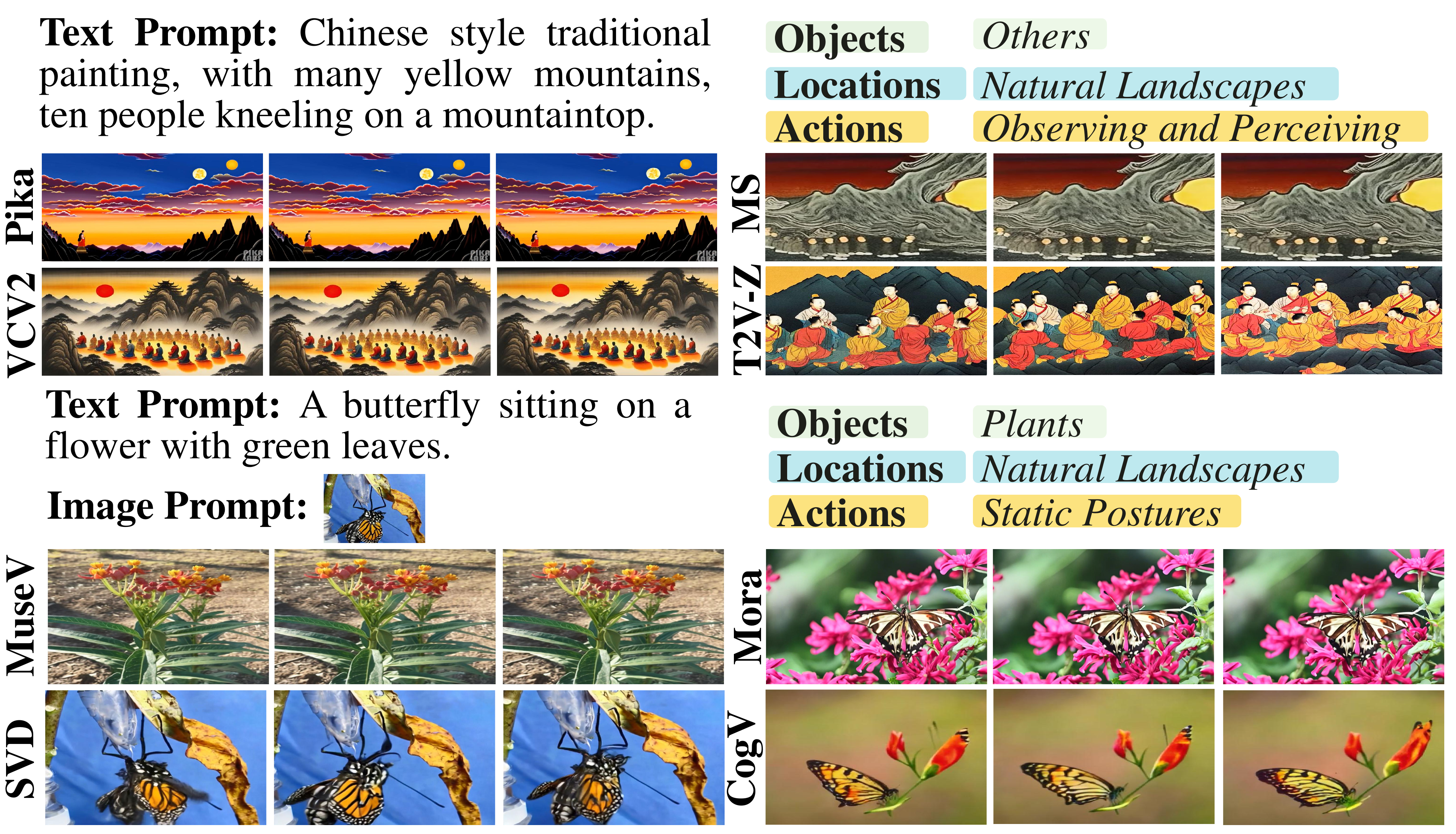}
    \caption{Overview of the proposed GenVidBench dataset. In addition to real/fake video labels, GenVidBench provides rich semantic annotations, including object categories, locations, and actions.
    } 
    \label{fig1} 
\end{figure}

\begin{table*}[ht]
    \centering
     \begin{tabular}{c|c|c|c|c|c}
     \toprule
           Dataset& \textbf{Scale} & \textbf{Prompt/Image} & \textbf{Video Pairs} & \textbf{Semantic Label} & \textbf{Cross Source} \\
     \midrule
       GVD~\cite{aigdet}   & 11k   & $\times$   & $\times$     & $\times$ & $\times$\\
       GVF~\cite{decof}   & 2.8k  & $\surd $   & $\surd $      & $\surd $  & $\times$\\
       GenVideo~\cite{demamba} & 2.27M & $\times$     & $\times$    & $\times$ & $\times$\\
       GenVidDet~\cite{genviddet} & 2.66M & $\times$     & $\times$    & $\times$ & $\times$\\
       \midrule 
       GenVidBench & 6.78M  & $\surd $      & $\surd $      & $\surd $ & $\surd $\\
     \bottomrule
     \end{tabular}%
     \caption{An overview of fake video detection datasets. The proposed GenVidBench is the first dataset with a scale of 6 million containing semantic labels and prompts/images used to generate videos. Furthermore, GenVidBench performs cross-source setting of the training set and the test set, which is extremely challenging for AI-generated video detection.}
    \label{related_work}%
\end{table*}%

GenVidBench is the first 6-million dataset for AI-generated video detection, and it features cross-source and cross-generator characteristics.
It covers 11 types of videos generated by state-of-the-art video generators, such as Mora, MuseV, Sora, and Kling, ensuring that the generated videos are of exceptional quality. GenVidBench contains two real video sources: Vript and HD-VG-130M. To make the dataset challenging, we construct two sets of paired videos, each derived from the same text prompt or image. The first video pair contains Pika, VideoCraftV2, Modelscope, and T2V-zero, which are included in the training set. The second video pair, which contains HD-VG-130M, MuseV, SVD, Mora, CogVideo, is included in the test set. For the second video pair, we generate prompts and images based on the real video and use Image-to-Video (I2V) models and Text-to-Video (T2V) models to generate videos from the same source. Videos with the same generation source have more similar attributes, which makes it highly difficult to distinguish real videos from AI-generated videos. In GenVidBench, the two sets of paired videos are organized into the training set and the test set, respectively, to make the dataset more challenging. Furthermore, the generators in the training and test sets are different, preventing videos generated by the same generator from sharing identical attributes. Based on this design, the task of cross-source and cross-generator detection is extremely challenging.

GenVidBench also provides rich semantic content labels for a portion of the videos. Researchers can select corresponding types of videos as required to adapt to different application scenarios. We describe the content of the video in three dimensions: object category, action, and location, which is shown in Figure~\ref{fig1}. 1) Object categories: the main character of the given video decides the major content. 2) Actions: a reflection of the temporal attributes. 3) Locations: an indicator of scenario complexity. Through analysis, we find that the object category affects the characteristics of the generated video. Through category differentiation, researchers are able to precisely identify and concentrate on the scenes of interest.

Comprehensive experimental results are presented to establish a solid foundation for researchers working on the development and assessment of AI-generated video detectors. A number of state-of-the-art video classification models are evaluated on the GenVidBench dataset, including DeMamba, VideoSwin, UniformerV2, etc. The findings from our experiments underscore the considerable challenge inherent in cross-source and cross-generator tasks. Moreover, we conduct a detailed analysis of particularly difficult cases, employing semantic content labels to identify and filter the most demanding categories. The results from these categories are presented in detail, offering an expanded set of benchmarks to further assist researchers in the field.

\begin{table*}[tbp]
    \centering
      \begin{tabular}{cccccccc}
      \toprule
      \textbf{Video Source} & \textbf{Train/Test} & \textbf{Type} & \textbf{Task} & \textbf{Resulution} & \textbf{FPS} & \textbf{V-6M} & \textbf{V-143k} \\
      \midrule
      Vript~\cite{vript} & Train & Real  & -     & -     & 30    & 417566 & 20131 \\
      Pika~\cite{pika}  & Train & Fake  & T2V\&I2V & 1088*560 & 24    & 1670465 & 13501 \\
      VideoCraftV2~\cite{vc2} & Train & Fake  & T2V\&I2V & 512*320 & 10    & 1672242 & 13501 \\
      Modelscope~\cite{modelscope} & Train & Fake  & T2V   & 256*256 & 8     & 1672242 & 13501 \\
      T2V-Zero~\cite{t2vz} & Train & Fake  & T2V   & 512*512 & 4     & 1268595 & 13501 \\
      OpenSora~\cite{open-sora} & Train & Fake  & T2V   & 256*256 & 8     & 13800 & 0 \\
      \hline
      HD-VG-130M~\cite{hd_vg_130m} & Test  & Real  & -     & 1280*720 & 30    & 13853 & 13853 \\
      MuseV~\cite{musev} & Test  & Fake  & I2V   & 1210*576 & 12    & 13853 & 13853 \\
      SVD~\cite{svd}   & Test  & Fake  & I2V   & 1024*576 & 10    & 13853 & 13853 \\
      Mora~\cite{mora}  & Test  & Fake  & T2V   & 1024*576 & 10    & 13853 & 13853 \\
      CogVideo~\cite{cogvideo} & Test  & Fake  & T2V   & 480*480 & 4     & 13853 & 13853 \\
      Sora~\cite{sora}  & Test  & Fake  & T2V   & 1920*1080 & 30    & 51    &  0\\
      Kling~\cite{keling} & Test  & Fake  & T2V\&I2V & -     & 30    & 264   &  0\\
      \midrule
      Sum   & -     & -     & -     & -     & -     & 6784490 & 143400 \\
      \bottomrule
      \end{tabular}%
      \caption{Statistics of real and generated videos in the GenVidBench dataset. GenVidBench contains 11 subsets of fake videos generated by 11 state-of-the-art generators and 2 subsets of real videos. }
    \label{datasets}%
  \end{table*}%
\section{Related Work}
\subsection{AI-Generated Content Detection Dataset}
With the rapid development of generative models, the requisite expertise and effort to generate content have been reduced. This has led to a growing focus on distinguishing real items from AI-generated content and constructing corresponding detection datasets. We categorize these detection datasets into three categories: AI-generated images, deepfake videos, and AI-generated videos.

\paragraph{AI-Generated Image Detection Dataset.}
As a result of blossoming of diffusion models, AI-generated images also have become more realistic.
GenImage~\cite{genimage} is a million-scale benchmark for detecting AI-generated images, which contains generated image pairs based on ImageNet~\cite{imgnet} using various diffusion-based models and GAN-based models.
Additionally, WildFake\cite{hong2024wildfake}, ArtiFact~\cite{rahman2023artifact}, and DiffusionDB~\cite{wang2023diffusiondb} also demonstrate the potential to provide a more comprehensive benchmark for fake image detection.

\paragraph{Deepfake Video Detection Dataset.}
A significant amount of research has focused on detecting deepfake videos~\cite{ju2023improving, Zhang_Yi_Wang_Zhang_Zeng_Tao_2024, zhao2021multiattentional, Ciamarra_2024_WACV, Hou_2023_CVPR}, based on deepfake datasets such as the Deepfake Detection Challenge Dataset (DFDC)~\cite{dolhansky2020deepfake}, Celeb-DF~\cite{li2020celebdf}, FaceForensics++~\cite{faceforensics}, and DeepFakeDetection (DFD)~\cite{dfd2019}. These datasets mainly use GANs, VAEs, or other swapping techniques to create fake videos. However, these datasets primarily consist of facial data and therefore lack diversity in content.

\paragraph{AI-Generated Video Detection Dataset.}
In the past few years, only a few studies have focused on detecting AI-generated videos across diverse scenarios. Generated Video Dataset (GVD)~\cite{aigdet} is constructed from 11,618 video samples produced by 11 generative models. GenVideo~\cite{demamba} is a large-scale AI-generated video detection dataset that collects videos from 10 different generator models for training and another 10 different generator models for testing. However, neither GVD nor GenVideo have the original prompts or images, video pairs, semantic labels, and cross-source settings, as shown in Table~\ref{related_work}. As a result, these datasets cannot avoid the problem of similar content between the training and test sets, and they are no way to distinguish between different scenarios. Generated Video Forensics (GVF)~\cite{decof} consists of video pairs from four different text-to-video models using the same prompts extracted from real videos. As illustrated in Table~\ref{related_work}, GVF contains prompts/images, video pairs, and semantic labels. However, it is too small, with only 2.8k videos, and does not include a cross-source setting. The proposed GenVidBench is the first dataset at the scale of 6.78 million videos, containing semantic labels and prompts/images used to generate videos. Furthermore, GenVidBench adopts a cross-source and cross-generator setting between the training and test sets, making fake video detection extremely challenging.

\subsection{Generated Video Detections}
AI-generated videos have the potential to accelerate the dissemination of misinformation, prompting significant concern. In the past, much of the research has concentrated on detecting videos with generated faces~\cite{tall}.
However, the content of the fake face video is single, which is greatly different from the real-world scenario. 
Due to the lack of large-scale high-quality AI-generated video datasets, there are few work on AI-generated video detectors. Recent work on AI-generated video detection includes AIGDet~\cite{aigdet}, DeCoF~\cite{decof}. 
AIGDet~\cite{aigdet} captures the forensic traces with a two-branch spatio-temporal convolutional neural network to improve detection accuracy. DeCoF~\cite{decof} is based on the principle of video frame consistency to eliminate the impact of spatial artifacts. Beyond the specialized models dedicated to the detection of generated videos, several video classification models have demonstrated remarkable performance on this task, including VideoSwin~\cite{videoswin} and UniFormerV2~\cite{uniformerv2}.
In this paper, we present the experimental results across various models on GenVidBench, which will provide a solid research foundation for developers in related fields. 

\begin{table*}[t]
    \centering
    \scalebox{1}{
      \begin{tabular}{c|c|c|c|c|c|c|c|c|c}
      \toprule
     \multirow{2}[4]{*}{\textbf{Training Set}} & \multicolumn{8}{c|}{\textbf{Test Set}}                        & \multicolumn{1}{c}{\multirow{2}[4]{*}{\textbf{Top-1}}} \\
   \cmidrule{2-9}          & \textbf{Pika} & \textbf{VC2} & \textbf{MS} & \textbf{T2VZ} & \textbf{MuseV} & \textbf{SVD} & \textbf{Cogvideo} & \textbf{Mora} &  \\
      \midrule
      Pika  & \textbf{96.93 } & 95.14  & 54.91  & 55.17  & 65.63  & 54.66  & 69.86  & 53.63  & 68.24  \\
      VC V2 & 66.75  & \textbf{99.52 } & 84.37  & 73.03  & 66.09  & 54.86  & 59.36  & 69.66  & 71.71  \\
      ModelScope    & 50.60  & 51.11  & \textbf{99.96 } & 52.68  & 59.68  & 54.37  & 41.03  & 49.11  & 57.32  \\
      T2V-Zero  & 50.56  & 51.98  & 55.46  & \textbf{99.96 } & 60.28  & 55.06  & 43.43  & 50.66  & 58.42  \\
      MuseV & 70.59  & 66.34  & 54.06  & 55.80  & \textbf{99.64 } & 62.55  & 73.54  & 53.14  & 66.96  \\
      SVD   & 50.60  & 61.63  & 62.37  & 65.99  & 94.27  & \textbf{99.71 } & 80.67  & 81.58  & 74.60  \\
      Cogvideo & 51.07  & 60.84  & 60.36  & 68.07  & 60.64  & 57.61  & \textbf{97.40 } & 52.55  & 63.57  \\
      Mora  & 51.84  & 76.14  & 62.12  & 62.48  & 60.96  & 86.17  & 51.84  & \textbf{99.42 } & 68.87  \\
      \bottomrule
      \end{tabular}
      }
      \caption{Results of cross-validation on different training and testing subsets using VideoSwin-Tiny.}
    \label{cross-val}%
\end{table*}%

\begin{figure*}[t]
    \centering
    \includegraphics[width=0.85\textwidth]{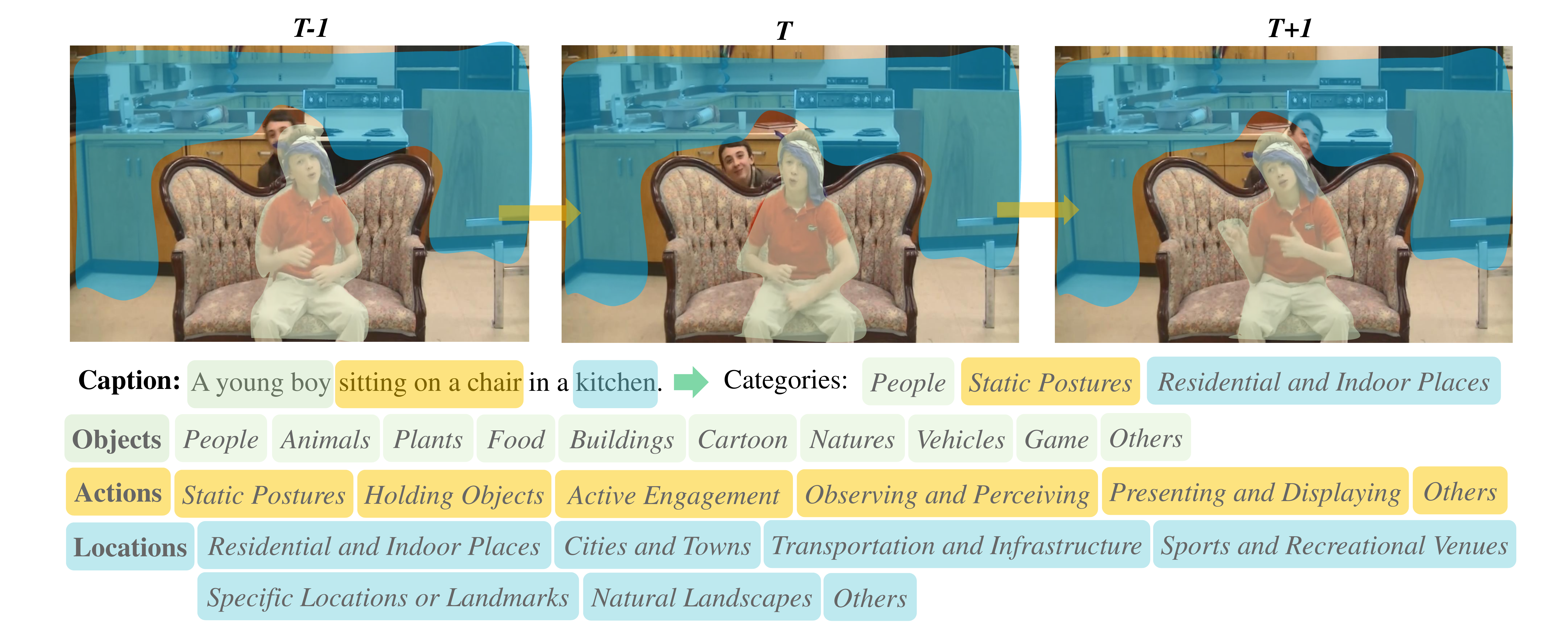}
    \caption{Semantic categorization of prompts in GenVidBench, including three dimensions: object categories, actions, and locations.}
    \label{caption}
\end{figure*}

\begin{table}[t]
    \centering
   \scalebox{0.88}{
      \begin{tabular}{c|c|c|c|c|c}
      \toprule
      \textbf{Pair1} & \textbf{Same} & \textbf{Different} & \textbf{Pair2} & \textbf{Same} & \textbf{Different} \\
      \midrule
      Pika  & \textbf{68.41 } & 60.95  & Musev & \textbf{63.08 } & 61.70  \\
      VC2   & \textbf{74.72 } & 62.49  & SVD  & \textbf{85.51 } & 60.15  \\
      MS   & \textbf{51.46 } & 51.05  & Cogvideo & 56.93  & \textbf{60.09 } \\
      T2V-Z  & \textbf{52.67 } & 52.36  & Mora  & \textbf{65.66} & 63.15  \\
      \midrule
      Mean  & \textbf{61.81 } & 56.71  & Mean  & \textbf{67.79 } & 61.27  \\
      \hline
   \end{tabular}
   }
   \caption{Comparison of VideoSwin-Tiny performance between training and test sets under same-source and cross-source generation. Results indicate that testing on different generation sources poses greater challenges.}
   \label{compare_source}
   \end{table}

\section{Dataset Construction}

\subsection{Overview of GenVidBench}
The GenVidBench dataset is distinguished by its large scale and exceptional diversity, providing a robust foundation for the development and evaluation of AI-generated video detection models. As illustrated in Table~\ref{datasets}, the dataset comprises a total of 6784490 video samples. This extensive scale ensures statistical robustness and facilitates the generalization of detection models across diverse scenarios, addressing the complexities inherent in real-world applications.

The AI-generated videos in the GenVidBench dataset are categorized into two main pairs, both derived from the same prompts or corresponding images. 
The first pair comprises videos from Pika, VideoCrafter2, Modelscope, and Text2Video-Zero. These videos are generated from a uniform prompt and sourced from the VidProM dataset~\cite{vidprom}, with each model providing approximately 1.67 million videos. 
The second pair includes videos from SVD, MuseV, Mora, CogVideo, and HD-VG-130M. Videos from SVD and MuseV are generated by extracting frames from the HD-VG-130M dataset. Similarly, Mora and CogVideo videos are generated using prompts from the HD-VG-130M dataset. The semantic content of the five video subsets is the same, and each model contributes 13,853 videos to the dataset.
By ensuring that the content of paired videos is identical, GenVidBench effectively prevents AI video detectors from distinguishing between real and generated videos based solely on content, thereby increasing the dataset's challenge and promoting the development of more sophisticated detection methods.

Moreover, the dataset exhibits remarkable diversity in terms of video types, tasks, resolutions, and frame rates. It encompasses both real and AI-generated content, supporting tasks such as Text-to-Video (T2V) and Image-to-Video (I2V). The resolutions range from 256$\times$256 in lower-quality samples to high-definition 1920$\times$1080, while frame rates vary from 4 FPS (e.g., CogVideo) to 30 FPS (e.g., HD-VG-130M). This heterogeneity ensures comprehensive coverage of diverse content, making GenVidBench a critical benchmark for advancing the performance and generalizability of AI-generated video detection models.

Due to the substantial size of GenVidBench, the required computational resources and training time can be significant. To address this issue, we introduce GenVidBench-143k, a lightweight version of the dataset specifically designed to facilitate rapid model iteration and development. As illustrated in Table~\ref{datasets}, this 143k subset is meticulously sampled from the original 6.78 million videos, ensuring that it remains both representative and sufficiently diverse. By focusing on this smaller yet comprehensive subset, researchers can conduct in-depth analysis with significantly reduced computational demands and time overhead, while still maintaining the broad applicability.

\begin{figure*}[t]
    \centering
    \includegraphics[width=0.95\textwidth]{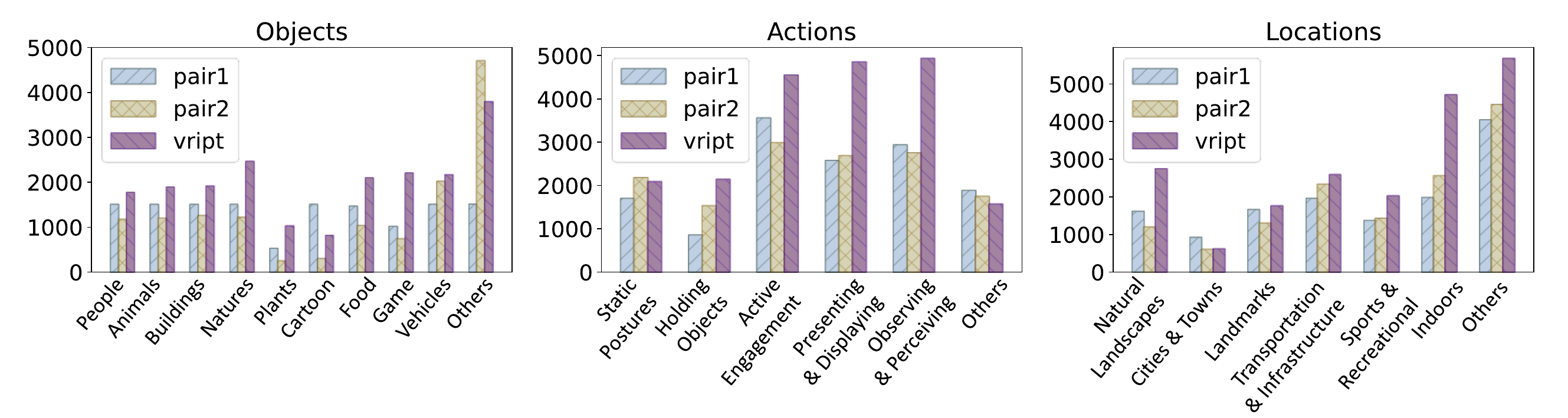}
    \caption{Data distribution across different video sources in GenVidBench-143k. Pair1 refers to Pika, VideoCrafterV2, Modelscope, and Text2Video-Zero. Pair2 refers to MuseV, SVD, CogVideo, Mora, and HD-VG-130M.}
    \label{content}
\end{figure*}

\begin{figure*}[t]
    \centering
    \includegraphics[width=0.94\textwidth]{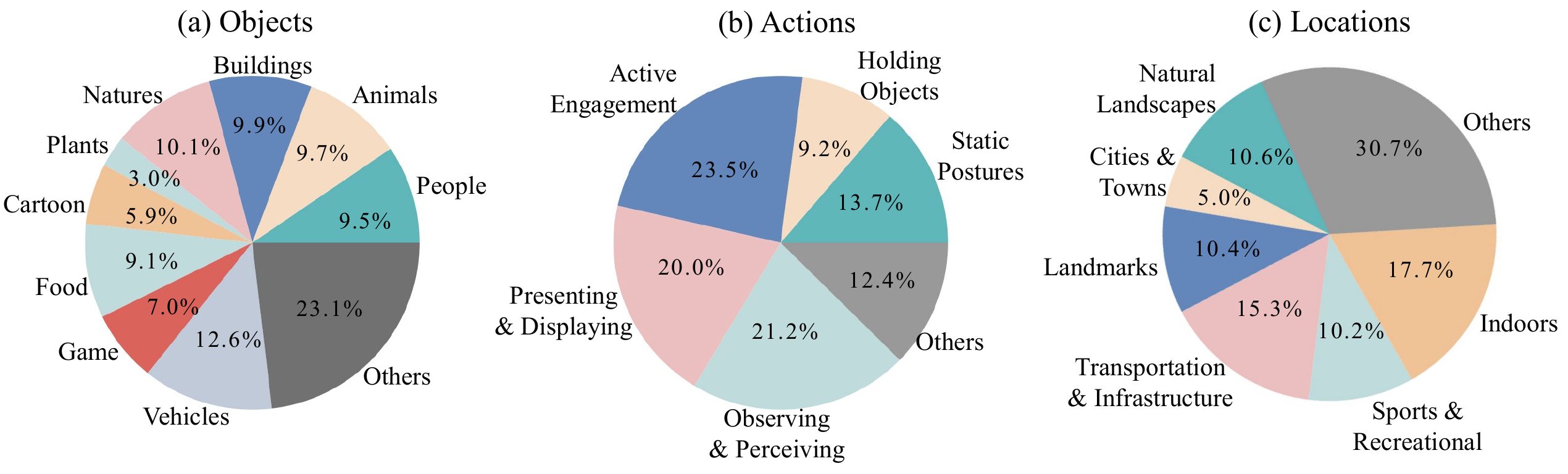}
    \caption{Data distribution across different semantic categories in GenVidBench-143k.}
    \label{all}
\end{figure*}

\subsection{Cross-Source and Cross-Generator Task}
GenVidBench contains two sets of video pairs with identical sources. The first video pair includes Pika, ModelScope, VideoCraftV2, and T2V-Zero, while the second video pair includes MuseV, SVD, CogVideo, Mora and HD-VG-130M. Each pair uses the same images or text for generation. To analyze the impact of different sources and different generators on the detection model, we carry out relevant evaluations.

First, we evaluate the performance of the detector when trained and tested on videos generated by the same generator. Our evaluation utilizes the widely recognized VideoSwin-tiny model~\cite{videoswin}. As shown in Table~\ref{cross-val}, training and testing within each subset consistently achieve accuracies above 97.40\%. Among them, the T2V-Zero and ModelScope subsets perform particularly well, with accuracies exceeding 99.90\%. These findings demonstrate that when the training and testing data are derived from the same generator, the task of detecting generated videos becomes relatively straightforward.

However, when testing on videos generated by a new generator, the experimental results are not satisfactory. It can be observed that a substantial performance degradation occurs when training and testing are conducted using different generators. For example, the accuracy of VideoSwin-Tiny drops to 54.66\% when trained on Pika and tested on SVD. In real-world applications, the identity of the generator is typically unknown, which poses a significant challenge for reliable detection.
To address this issue, our study aims to rigorously assess the generalization capability of detectors. Specifically, their ability to distinguish between real and generated videos without dependence on the generator’s identity. To this end, we introduce the cross-generator video classification task, aimed at evaluating the robustness and adaptability of detectors across diverse video sources.

Furthermore, we analyze the influence of the generation source, such as text prompts and images, on the classification of real and generated videos. The four models Pika, VideoCrafter2, Modelscope, and Text2Video-Zero use the same text prompt to generate a video, and these videos are the first video pair. The MuseV and SVD models are based on the same image for video generation, while CogVideo and Mora rely on the same text prompt, forming our second set of video pairs.
As shown in Table~\ref{compare_source}, training with videos from the same video source tends to yield superior test results. For example, in the first video pair, Pika achieves an accuracy of 68.41\% when tested on videos from the same source, but only 60.95\% when tested on videos generated by other sources. Similarly, in the second video pair, SVD reaches 85.51\% accuracy when trained and tested on the same source, yet this drops to 60.15\% when evaluated on videos from different sources. Moreover, across both pairs, the average accuracy obtained with same-source testing consistently exceeds that of cross-source testing. These findings highlight the strong dependency of detectors on the generation source.
Based on the above analysis, we propose a challenging dataset and introduce a cross-source and cross-generator task to enhance detector robustness and generalization. The videos in Video Pair 1 are used as the training set, while the videos in Video Pair 2 are used as the test set. This design prevents the model from distinguishing real and fake videos merely based on content or video quality, thereby increasing the difficulty of the task and better simulating real-world conditions.

\begin{table*}[tbp]
    \centering
    \scalebox{0.9}{
      \begin{tabular}{c|c|c|c|c|c|c|c|c|c}
      \toprule
      \textbf{Method} & \textbf{Type} & \textbf{MuseV} & \textbf{SVD} & \textbf{CogV} & \textbf{Mora} & \textbf{Sora} & \textbf{Kling} & \textbf{HD} & \textbf{Top-1} \\
      \midrule
      I3D~\cite{i3d}   & CNN   & 32.72 & 12.04 & 76.44 & 72.3  & 41.18 & 46.22 & 95.04 & 60.21 \\
      SlowFast~\cite{slowfast} & CNN   & 87.14 & 29.80  & 93.07 & 55.23 & 23.53 & 58.33 & 96.61 & 70.06 \\
      TPN~\cite{tpn}   & CNN   & 56.34 & 24.30  & 71.87 & 73.37 & 9.80   & 46.59 & 99.07 & 66.23 \\
      TIN~\cite{tin}   & CNN   & 68.38 & 41.44 & 57.88 & 79.67 & 35.29 & 60.61 & 96.41 & 67.91 \\
      TRN~\cite{trn}   & CNN   & 43.52 & 27.16 & 58.30  & 97.26 & 29.42 & 58.72 & 97.56 & 66.67 \\
      TSM~\cite{tsm}  & CNN   & \textbf{95.94} & \textbf{73.16} & 36.44 & 91.72 & 33.34 & \textbf{71.6} & 96.30  & 73.88 \\
      \midrule
      TimeSformer~\cite{timesformer} & Trans. & 69.94 & 29.13 & 67.82 & \textbf{98.47} & \textbf{64.71} & 65.91 & 72.47 & 71.53 \\
      UniFormerV2~\cite{uniformerv2} & Trans. & 27.97 & 12.84 & 94.79 & 78.21 & 5.88  & 21.59 & 99.46 & 65.31 \\
      VideoSwin~\cite{videoswin} & Trans. & 90.24 & 27.72 & 91.64 & 88.14 & 19.60  & 50.76 & 99.10  & 80.39 \\
      MViTv2-S~\cite{mvit} & Trans. & 77.08 & 44.89 & \textbf{99.91} & 76.77 & 61.36 & 31.37 & 86.98 & 80.45 \\
      \midrule
      DeMamba~\cite{demamba} & Mamba & 85.04 & 48.81 & 98.66 & 90.23 & 1.96  & 33.71 & \textbf{99.86} & \textbf{85.47} \\
      \bottomrule
      \end{tabular}%
    }
      \caption{Results of various state-of-the-art methods trained on cross-source and cross-generator task.}
    \label{results}%
\end{table*}%

\subsection{Content Analysis}
Taking the 143k dataset as a representative example, we analyze the content distribution of the GenVidBench dataset.
To facilitate the development of more generalized and effective detection models, we categorize prompts across multiple dimensions.
As depicted in Figure~\ref{caption}, we initially divide prompts into three key dimensions: object categories, actions, and locations, as these three elements can form a complete sentence or story with rich semantics.

1) \textbf{Object categories}. The main subject of the given video determines the primary spatial content.
2) \textbf{Actions}. Reflect the temporal attributes of the video.
For instance, categorizing the action as Static Postures implies that the main subject of the video will not exhibit rapid movement.
If the action is categorized as Presenting and Displaying, the main subject may engage in walking around objects or performing other bodily movements.
3) \textbf{Locations}.
This indicates the complexity of the scenario. For example, a Nature Landscape may have a clean background, whereas a Transportation scenario might include intricate machinery in the background.

For each dimension, taking the object category dimension as an example, we sample the entire prompt set and utilize Large Language Models (LLMs) to extract the subjects. We then aggregate these subjects into more abstract categories. The final categories are constrained to no more than ten options to construct a classification tree. For object dimension, we use people, animals, buildings, natures, plants, cartoon, food, game, vehicles, others as our final classification tree. Based on the classification tree, we can obtain the class labels for each prompt of our entire dataset.

We select object categories as our primary dimension for selecting and generating video pairs, ensuring that our benchmark is balanced and semantically rich in the spatial dimension. Figure~\ref{content} illustrates the distribution of various prompt sets across multiple dimensions. Figure~\ref{all} presents the distribution across different semantic categories. These figures demonstrate that our benchmark encompasses a wide variety of video content, spanning broad category distributions across both spatial and temporal dimensions.

\begin{table}[tbp]
    \centering
    \scalebox{0.89}{
      \begin{tabular}{c|c|c|c|c}
      \toprule
      \textbf{Method } & \textbf{Type} & \textbf{Top-1} & \textbf{F1} & \textbf{AUROC } \\
      \midrule
      I3D   & CNN   & 60.21 & 67.38 & 84.06 \\
      SlowFast & CNN   & 70.06 & 81.13 & 88.65 \\
      TPN   & CNN   & 66.23 & 73.39 & 95.67 \\
      TIN   & CNN   & 67.91 & 73.74 & 72.96 \\
      TRN   & CNN   & 66.67 & 72.49 & 92.13 \\
      TSM   & CNN   & 73.88 & 82.49 & 95.31 \\
      \midrule
      TimeSformer & Transformer & 71.53 & 77.31 & 76.71 \\
      UniFormerV2 & Transformer & 65.31 & 73.03 & 96.73 \\
      VideoSwin & Transformer & 80.39 & 86.36 & 96.05 \\
      MViTv2 & Transformer & 80.45 & 85.66 & 90.29 \\
      \midrule
      DeMamba & Mamba & \textbf{85.47} & \textbf{90.27} & \textbf{99.28} \\
      \bottomrule
      \end{tabular}%
    }
      \caption{Comparison of Top-1 accuracy, F1 score, and AUROC across different methods.}
    \label{f1_auroc}%
\end{table}%

\subsection{Scenario-Specific Task}
As mentioned above, we conduct a comprehensive content analysis of all prompts and videos to obtain category labels across multiple dimensions.
This process enables researchers to extract datasets customized to their specific research interests.
For instance, they can extract scenes of People to evaluate the logic of human body generation, scenes of Presenting to study motion deformation, and scenes with low temporal continuity to analyze frame-wise deformation.
Furthermore, based on the extracted dataset, we can evaluate the impact of these attributes on the detector, such as the relationship between motion deformation and detection accuracy.

\begin{table}[tbp]
    \centering
    \scalebox{0.8}{
        \begin{tabular}{c|c|c|c|c|c|c}
        \toprule
        \textbf{Method} & \textbf{MuseV} & \textbf{SVD} & \textbf{CogV} & \textbf{Mora} & \textbf{HD} & \textbf{Top-1} \\
    \midrule
    I3D   & 8.15  & 8.29  & 60.11  & 59.24  & 93.99  & 45.96  \\
    SlowFast & 12.25  & 12.68  & 38.34  & 45.93  & 93.63  & 40.57  \\
    TPN   & 37.86  & 8.79  & 68.25  & 90.04  & 97.34  & 60.46  \\
    TIN & 33.78  & 21.47  & 81.59  & 79.44  & 97.88  & 62.83  \\
    X3D & \textbf{92.39} & 37.27  & 65.72  & 49.60  & 97.51  & 68.50  \\
    TRN & 38.92  & 26.64  & \textbf{91.34} & 93.98  & 93.97  & 68.97  \\
    TSM & 70.37  & 54.70  & 78.46  & 70.37  & 96.76  & 74.13  \\
    \midrule
    UniFormerV2 & 20.05  & 14.81  & 45.21  & \textbf{99.21} & 96.89  & 55.23  \\
    TimeSformer & 73.14  & 20.17  & 74.80  & 39.40  & 92.32  & 59.97  \\
    VideoSwin & 49.11  & 68.48  & 88.47  & 81.42  & \textbf{98.55} & 77.21  \\
    MViTv2-B & 76.34  & \textbf{98.29} & 47.50  & 96.62  & 97.58  & \textbf{83.27} \\
        \bottomrule
        \end{tabular}%
    }
        \caption{The results in GenVidBench-143k.}
      \label{143k}%
\end{table}

\begin{table}[tbp]
    \centering
    \scalebox{0.9}{
    \begin{tabular}{c|c|c|c}
        \toprule
        \textbf{Dataset} & \textbf{SlowFast} & \textbf{I3D} & \textbf{F3Net} \\
        \midrule
        DeepFakes~\cite{dfd2019} & 97.53 & -     & - \\
        Face2Face~\cite{face2face}  & 94.93 & -     & - \\
        FaceSwap~\cite{faceswap}  & 95.01 & -     & - \\
        NeuralTextures~\cite{thies2019deferred}  & 82.55 & - & -\\ %
        GVF~\cite{decof} & - & 61.88  & - \\
        GenVideo~\cite{demamba} & -     & - & 51.83  \\
        \hline
        GenVidBench(Ours) & \textbf{70.06} & \textbf{60.21}  & \textbf{42.52}  \\
        \bottomrule
        \end{tabular}
        }
        \caption{Performance comparison of detectors on different datasets.}
        \label{dataset_compare}%
\end{table}

\begin{table*}[ht]
    \centering
    \scalebox{0.95}{
      \begin{tabular}{c|c|c|c|c|c|c|c|c|c|c}
      \toprule
      \textbf{Method} & \textbf{People} & \textbf{Animals} & \textbf{Buildings} & \textbf{Natures} & \textbf{Plants} & \textbf{Cartoon} & \textbf{Food}  & \textbf{Game} & \textbf{Vehicles} & \textbf{Top-1} \\
      \midrule
      SVD & 0.757 & 0.728 & 0.777 & 0.736 & 0.739 & 0.537 & 0.769 & 0.732 & \textbf{0.795} & 0.739 \\
      MuseV & 0.156 & 0.152 & 0.16 & 0.164 & \textbf{0.220} & 0.098 & 0.117 & 0.189 & 0.129 & 0.160 \\
      Mora & \textbf{0.226} & 0.153 & 0.196 & 0.210 & 0.205 & 0.157 & 0.171 & 0.156 & 0.215 & 0.198 \\
      CogVideo & 0.014 & 0.008 & 0.010 & 0.023 & 0.012 & 0.004 & 0.010 & \textbf{0.036} & 0.013 & 0.017 \\
      \midrule
      Mean & 0.307 & 0.277 & 0.306 & 0.303 & \textbf{0.318} & 0.209 & 0.286 & 0.295 & 0.308 & 0.297 \\
      \bottomrule
      \end{tabular}}%
       \caption{Distribution of hard cases across categories in video subsets generated by different generators.}
    \label{tab:hardcase}%
   \end{table*}%

\begin{table}[t]
    \centering
    \scalebox{0.8}{
     \begin{tabular}{c|c|c|c|c|c|c}
     \toprule
     \textbf{Method} & \textbf{MuseV} & \textbf{SVD} & \textbf{CogV} & \textbf{Mora} & \textbf{HD} & \textbf{Mean} \\
     \midrule
     I3D  & 39.18  & 23.27  & 91.98  & 78.38  & 78.42  & 62.15  \\
     SlowFast & 81.63  & 29.80  & 75.31  & 19.31  & 73.03  & 55.30  \\
     TPN  & 43.67  & 20.00  & 85.80  & 86.87  & 94.61  & 64.24  \\
     TimeSformer& 77.96  & 29.80  & 96.30  & 93.44  & 87.14  & 75.09  \\
     VideoSwin & 57.96  & 7.35  & 92.59  & 47.88  & 98.76  & 52.86  \\
     UniFormerV2 & 13.88  & 7.76  & 41.98  & 95.75  & 97.93  & 64.76  \\
     \bottomrule
     \end{tabular}
    }
    \caption{Results of various state-of-the-art methods trained on plants class. 
    }
    \label{plants_result}%
\end{table}%

\section{Experimental Analysis}

\subsection{Implementation Details}
We sample 8 frames from each video as input and resize each image to 224$\times$224. The batch size is 8. We use the default learning rate provided by MMAction2~\cite{mmaction2} for each method. We also apply data augmentation methods such as random flipping and cropping. All other training settings follow the default configuration of MMAction2. 

\subsection{Results of Cross-Source and Cross-generator Task}
To establish a reliable benchmark, we evaluate the performance of multiple methods on GenVidBench-6M. As shown in Table~\ref{results}, DeMamba achieves the highest accuracy with a Top-1 of 85.47\%, followed by MViTv2 with 80.45\%. Overall, Transformer-based models outperform CNN-based models. Among CNNs, TSM achieves the best performance with a Top-1 accuracy of 73.88\%, which is 6.57\% lower than MViTv2. These results demonstrate that the cross-source and cross-generator task is highly challenging, leaving substantial room for improvement in generative video detection. Furthermore, real videos are more easily distinguished from fake ones, with most classification accuracies exceeding 95.04\%. Fake videos generated by Sora tend to be more challenging to classify correctly, reflecting their superior generation quality, while those produced by CogVideo are generally easier to detect due to limited temporal continuity. F1 and AUROC are also reported in Table~\ref{f1_auroc}.

The experimental results on GenVidBench-143k are presented in Table~\ref{143k}. MViTv2 still achieves a high accuracy. However, owing to the smaller training set, the overall accuracy is lower than that obtained on GenVidBench-6m.

\subsection{Comparison with Advanced Datasets}
To compare the challenges posed by different datasets, we conduct a comparative analysis of the performance of SlowFast, I3D, and F3Net across a range of datasets, as shown in Table~\ref{dataset_compare}. Following the results reported in previous work~\cite{decof,demamba}, we additionally include new results trained on our benchmark. It can be observed that SlowFast, I3D, and F3Net achieve significantly better performance on other datasets than on GenVidBench. For instance, SlowFast reaches 82.55\% accuracy on the NeuralTextures dataset, but only 70.06\% on GenVidBench. Similarly, I3D achieves 61.88\% accuracy on the GVF dataset, compared with 60.21\% on GenVidBench. These results demonstrate that our dataset is more challenging.

\subsection{Hard Case Analysis}
To further investigate the detection results, we conduct a hard-case analysis based on the experimental outcomes of VideoSwin-tiny. We define hard cases as generated videos for which the detector assigns a very low probability. The results are summarized in Table~\ref{tab:hardcase}. While the detector demonstrates strong and consistent performance across all test videos, its effectiveness varies considerably in cross-generator scenarios. As shown in Table~\ref{tab:hardcase}, SVD poses more challenging cases for the detector, whereas CogVideo produces almost no severely ambiguous cases.

\subsection{Results on Scenario-Specific Task} 
Based on the analysis of hard examples, we select the Plants class and give the experimental results of various models on this class. As illustrated in Table~\ref{plants_result}, TimeSformer achieves the best accuracy, with a Top-1 of 75.09\%. Additionally, TPN and UniFormerV2 also demonstrate commendable classification esults, with Top-1 accuracies of 64.24\% and 64.76\%, respectively. Conversely, SlowFast lags behind with a Top-1 accuracy of only 55.30\%. VideoSwin achieves the poorest performance, attaining a mere 52.86\% in Top-1 accuracy. Experimental results also show that SVD has the lowest classification accuracy, which proves that its generation performance is the best. 
The above experimental results show that the classification performance of the model in a single scene is different from that in all scenes, so it is necessary to detection for different scenes. The rich semantic labels provided by GenVidBench can help analyze each scenario and provide more development ideas for developers.

\section{Conclusion}
In this paper, we introduce GenVidBench, a large-scale dataset comprising six million videos for AI-generated video detection. The dataset is characterized by its cross-source and cross-generator setup, and it provides diverse semantic content labels. We present comprehensive experimental results to establish a solid foundation for future research on the development and evaluation of AI-generated video detectors. The experiments demonstrate that GenVidBench is more challenging than existing datasets, thereby offering greater opportunities for advancing detectors.

\bibliography{main}

\end{document}